\documentclass{article}

\usepackage{PRIMEarxiv}

\usepackage[utf8]{inputenc} % allow utf-8 input
\usepackage[T1]{fontenc}    % use 8-bit T1 fonts
\usepackage{hyperref}       % hyperlinks
\usepackage{url}            % simple URL typesetting
\usepackage{booktabs}       % professional-quality tables
\usepackage{amsfonts}       % blackboard math symbols
\usepackage{nicefrac}       % compact symbols for 1/2, etc.
\usepackage{microtype}      % microtypography
\usepackage{lipsum}
\usepackage{fancyhdr}       % header
\usepackage{graphicx}       % graphics

\usepackage{multirow}%
\usepackage{amsmath,amssymb,amsfonts}%
\usepackage{amsthm}%
\usepackage{mathrsfs}%
\usepackage[title]{appendix}%
\usepackage{xcolor}%
\usepackage{textcomp}%
\usepackage{manyfoot}%
\usepackage{booktabs}%
\usepackage{algorithm}%
\usepackage{algorithmicx}%
\usepackage{algpseudocode}%
\usepackage{listings}%
\usepackage{verbatim}%
\usepackage[normalem]{ulem}%

\graphicspath{{media/}}     % organize your images and other figures under media/ folder

\title{MindfulLIME: A Stable Solution for Explanations of Machine Learning Models with Enhanced Localization Precision - A Medical Image Case Study
}

\author{
  Shakiba Rahimiaghdam \\
  Department of Computer Engineering \\
  Middle East Technical University \\
  Ankara, Turkey, 06800\\
  \texttt{shakiba.rahimiaghdam@metu.edu.tr} \\
   \And
  Hande Alemdar \\
  Department of Computer Engineering \\
  Middle East Technical University \\
  Ankara, Turkey, 06800\\
  \texttt{alemdar@metu.edu.tr} \\
}

\begin{document}
\maketitle

\begin{abstract}
Ensuring transparency in machine learning decisions is critically important, especially in sensitive sectors such as healthcare, finance, and justice. Despite this, some popular explainable algorithms, such as Local Interpretable Model-agnostic Explanations (LIME), often produce unstable explanations due to the random generation of perturbed samples. Random perturbation introduces small changes or noise to modified instances of the original data, leading to inconsistent explanations. Even slight variations in the generated samples significantly affect the explanations provided by such models, undermining trust and hindering the adoption of interpretable models. To address this challenge, we propose MindfulLIME, a novel algorithm that intelligently generates purposive samples using a graph-based pruning algorithm and uncertainty sampling. MindfulLIME substantially improves the consistency of visual explanations compared to random sampling approaches. Our experimental evaluation, conducted on a widely recognized chest X-ray dataset, confirms MindfulLIME's stability with a 100\% success rate in delivering reliable explanations under identical conditions. Additionally, MindfulLIME improves the localization precision of visual explanations by reducing the distance between the generated explanations and the actual local annotations compared to LIME. We also performed comprehensive experiments considering various segmentation algorithms and sample numbers, focusing on stability, quality, and efficiency. The results demonstrate the outstanding performance of MindfulLIME across different segmentation settings, generating fewer high-quality samples within a reasonable processing time. By addressing the stability limitations of LIME in image data, MindfulLIME enhances the trustworthiness and interpretability of machine learning models in specific medical imaging applications, a critical domain.
\end{abstract}

% keywords can be removed
\keywords{Explainable Artificial Intelligence (XAI) \and Classification \and Deep Learning \and Neural Networks}

\section{Introduction}\label{section:1}
The advent of machine learning-based systems has revolutionized numerous complex tasks in our lives, ranging from image object detection and text processing to speech signal analysis \cite{DONG2021100379,doi:https://doi.org/10.1002/9781119861850.ch7}. These systems have demonstrated remarkable performance, often surpassing human-level capabilities. However, their inherent non-transparency, non-linearity, and nested internal computations pose a challenge in establishing user trust \cite{vonEschenbach2021}. Most machine learning algorithms lack traceability, resembling black boxes where decisions are made without clear explanations based on input data. Hence, it is crucial to have deep learning models that are explainable and transparent in order to enable their widespread utilization across different vertical domains, particularly in critical areas like justice, cybersecurity, and medicine \cite{COLLENETTE2023103861,https://doi.org/10.1002/qre.2939,https://doi.org/10.1002/wsbm.1548}.

Simpler algorithms, including decision trees and linear regression, offer greater transparency owing to their clear structures. In contrast, complex models, notably neural networks and random forests, tend to operate as black boxes. Addressing this, the field of Explainable Artificial Intelligence (XAI) endorses interpretable models to enhance transparency in such systems \cite{Arrieta2019}. These models are classified as either model-agnostic or model-specific, spanning from local to global, and intrinsic to post hoc explanations \cite{Linardatos2021}. Notably, post hoc model-agnostic approaches like Local Interpretable Model-agnostic Explanations (LIME) \cite{zhang2019why}, Anchors \cite{Ribeiro_Singh_Guestrin_2018}, and SHapley Additive exPlanations (SHAP) \cite{10.5555/3295222.3295230} are prevalent. They elucidate the decision-making processes of machine learning models, irrespective of internal model structures.

LIME, a prominent algorithm in XAI, leverages simpler models like logistic regression near the classification boundary to provide clear and understandable explanations. It decodes decision-making processes by generating simulated data points through perturbation and fitting a sparse linear model, thus can potentially offer explanations suitable for a variety of models and data types, including text, images, or tabular data. The explanations remain locally faithful, irrespective of the classifier used \cite{zhang2019why}. However, the stability of interpretative models is impacted due to LIME's reliance on random perturbation \cite{hooker2021unrestricted}. Such non-deterministic methods can lead to inconsistencies, data and label shifts, and overlook feature correlations, making it crucial to refine sampling operations to mitigate instability and prevent out-of-distribution data, especially in critical fields like medicine \cite{KHAIRE20221060,10.1016/j.inffus.2021.05.009,10.5555/3327757.3327875,rahnama2019study,Slack2020}.

This paper examines post hoc explanations with perturbations, focusing on image data, where stability is notably challenging. In response, we introduce MindfulLIME, a novel algorithm enhancing LIME with intelligent non-random sampling for improved stability and reproducibility. The designation MindfulLIME reflects our approach to augment LIME with careful sampling strategies, making the algorithm more cognizant of stability in its explanations. Our decision to center on medical images and specifically X-ray images in this study is driven by their pivotal role in medical diagnostics, the complexities and challenges tied to interpreting these images, and the accessibility of publicly available datasets for experimentation. This strategy not only accentuates the specific application of medical images addressed herein but also suggests the potential for MindfulLIME to be efficaciously generalized across other domains.

In our framework, \emph{explanation} refers to image superpixels or segments crucial to the classifier's decision. We enhance the stability of these visual explanations by transforming superpixels into an undirected graph, where each segment is a vertex. This graph structure facilitates systematic generation of new samples by deactivating superpixels in iterations. Each generated sample undergoes evaluation to determine its retention or removal, guided by uncertainty sampling based on the classifier’s confidence level. This technique prioritizes samples for which the classifier exhibits lower confidence (uncertainty sampling) \cite{10.1007/978-1-4471-2099-5_1} or higher confidence (reversed uncertainty sampling) \cite{Cardellino2015ReversingUS}, enabling us to bolster the quality and reliability of our explanations. By focusing on confident samples, we ensure their informativeness and accurate representation of data distribution, while minimizing out-of-distribution risks. Additionally, our method considers feature interplay to further enhance the quality of explanations.

In the sensitive medical domain, explainability is crucial, particularly for chest X-ray (CXR) image analysis in thorax disease diagnosis, where human interpretation can be limited and complex, increasing misdiagnosis risks \cite{FOURCADE2019279,VANDERVELDEN2022102470}. Our framework, leveraging the CheXpert \cite{article} and VinDr-CXR datasets \cite{Nguyen2020}, addresses these challenges. We developed MindfulLIME, an algorithm that enhances the interpretability and precision of AI-driven diagnostic tools. MindfulLIME, by systematically selecting superpixel combinations and employing non-random sampling, achieves a 100\% stability rate and surpasses LIME in localization precision. This approach not only improves the quality and reliability of explanations but also mitigates the risk of out-of-distribution samples, demonstrating its potential in enhancing medical professionals' decision-making capabilities and advancing medical research.

The remaining sections of the manuscript are structured as follows: Section~\ref{section:2} provides an overview of related studies in the literature focusing on improvements for perturbed-based sampling algorithms, including LIME. Section~\ref{section:3} presents our proposed approach in detail, preceded by a brief background on LIME. The experimental setup, including the utilized datasets and the trained classifier setting, is described in Section~\ref{section:4}. Section~\ref{section:5} presents the discussion and results of our extensive evaluations of the VinDr-CXR dataset. We compare MindfulLIME and LIME using various evaluation criteria based on different super-pixel algorithms and various sample numbers. Finally, we conclude our work with Section~\ref{section:6}.

\section{Related Works}\label{section:2}

Perturbation-based techniques have shown promise in explaining black-box machine learning models, but they have several issues that must be addressed. These include generating out-of-distribution data points, which can lead to less reliable classifier outputs and inconsistent results in LIME \cite{10.1007/978-3-030-69544-6_7}. Furthermore, these techniques often fail to consider useful information, such as feature correlations, when generating new samples \cite{Shi2020}. As a result, the explanations generated by these approaches are relatively less reliable, less trustworthy, and of poor quality. This section discusses relevant works that aim to improve LIME or similar models from various aspects. These approaches encompass various strategies, including identifying more suitable random samples, utilizing additional data, generating synthetic In-Distribution datasets, or completely altering the sample generation process to eliminate randomness. Indeed, it is worth noting that these approaches also address distinct issues.

Inspired by stable approximation trees in model distillation \cite{zhou2018approximation}, Zhou et al. \cite{10.1145/3447548.3467274} introduced S-LIME, an approach that tackles the challenge of stability in LIME and similar techniques. S-LIME leverages a hypothesis testing framework based on the central limit theorem to determine the optimal number of perturbed samples required to generate more stable explanations than traditional LIME. OptiLIME, another framework proposed by Visani et al. \cite{visani2022optilime}, focuses on maximizing stability and maintaining fidelity to ML model explanations. It achieves this by automatically determining the optimal kernel width value, ensuring the desired fidelity level while optimizing explanation stability. In contrast to our integrated solution, these techniques require individual execution for each dataset or problem, increasing the workload for each case. While they improve stability, they do not provide a guarantee of a 100\% stable solution. Additionally, the run-time efficiency and potential impact on the quality of explanations have yet to be thoroughly examined.

Some approaches generate datasets that preserve the distribution of the training data around the target instance, resulting in stable and accurate explanations. For example, authors \cite{Jiang2022} proposed a modified version of LIME for regression models in aerodynamic tabular data. Their approach involves perturbed sample generation that emphasizes differences and ensures a uniform distribution of distances and weights. However, further research is needed to validate this approach with more extensive and diverse datasets, different models, and additional analyses. A study \cite{vreš2021better} proposes advanced sampling methods utilizing data generators to capture the training dataset's distribution effectively. The authors explore various data generators, including RBF network-based, and variational autoencoder-based approaches \cite{7112148,8959787}. The focus is on robustness, which refers to the ability of an explanation method to identify biased classifiers in adversarial environments, mainly targeted attacks \cite{Slack2020, alvarezmelis2018robustness}. The research highlights the importance of data generators and emphasizes the need for requirements regarding access to training data. Similarly, Saito et al. \cite{Saito2020} introduce a method to enhance the robustness of LIME against adversarial attacks. They propose training a Conditional Tabular GAN (CTGAN) model \cite{10.5555/3454287.3454946}, a generative adversarial network, to generate realistic synthetic data for LIME's explanation generation process. Experimental results on three real-world datasets demonstrate improved accuracy in detecting adversarial behavior compared to vanilla LIME while maintaining comparable explanation quality.

Generating synthetic data demands particular attention to detail and significant effort, involving creating artificial data points that closely resemble the distribution and characteristics of the original training data. This can be challenging for high-dimensional or complex datasets, requiring computational resources, parameter tuning, and iterative refinement. Integrating synthetic data generation into the explanation process adds complexity and computational overhead, necessitating customized approaches and parameter settings for each dataset. Careful consideration of computational resources, time, and expertise is crucial, along with optimizing implementation to mitigate the additional load.

On the other hand, alternative methods take a different approach by utilizing a portion of the training data instead of generating a synthetic dataset. Hall et al. \cite{Hall2019} propose a technique that involves partitioning the dataset using k-means clustering instead of perturbing data points around the instance being explained. However, the non-deterministic nature of the default k-means implementation, which randomly selects centroids, limits the effectiveness of this approach. Similarly, Hu et al. \cite{hu2018locally} introduce a supervised tree-based method for dataset partitioning. Building upon these ideas, Zafar et al. \cite{Zafar2021} present a more practical approach called Deterministic Local Interpretable Model-Agnostic Explanations (DLIME) to enhance the interpretability of black box Machine Learning (ML) algorithms. DLIME utilizes Agglomerative Hierarchical Clustering (AHC) and K-Nearest Neighbor (KNN) techniques instead of generating synthetic samples, providing a deterministic variation of LIME. It is important to note that this approach requires access to the training data and may face challenges when applied to image data or datasets with a large number of features due to the inherent complexity involved.

Some approaches use methods to select relevant samples or modify random samples to improve their suitability. Applying criteria like affinity quantification or inlier scores in the filtering step enhances the quality and effectiveness of the explanations. For instance, a solution proposed by the authors \cite{Qiu2022} addresses the Out-of-Distribution problem by incorporating a module that measures the similarity between perturbed data and the original dataset distribution. The authors enhance the robustness of the final explanations by penalizing the influence of unreliable OoD data, leveraging inlier scores and prediction results. While their approach claims compatibility with popular perturbation-based XAI algorithms such as RISE \cite{Petsiuk2019}, OCCLUSION \cite{10.1007/978-3-319-10590-1_53}, and LIME, they acknowledge its inapplicability to LIME and identify deficiencies in localization metrics. Rasouli et al. \cite{Rasouli2019} address limitations in traditional data sampling for local explanation methods. They propose a sampling methodology based on observation-level feature importance, which generates more meaningful perturbed samples than Gaussian or training data distributions. The proposed approach captures feature interactions and creates a perturbed dataset that accurately represents the locality of a given sample. However, further testing is needed to evaluate its stability and efficiency, and its applicability is currently limited to tabular data. ALIME \cite{10.1007/978-3-030-33607-3_49} utilizes a pre-trained autoencoder to transform randomly generated disturbance samples into lower-dimensional latent codes. Instead of relying on Euclidean distance, the similarity between disturbed samples and the input instance is assessed based on the distance between these latent codes. This approach offers an alternative way to measure similarity and capture significant patterns during the explanation process.

In contrast to random perturbation, deterministic perturbation techniques have been introduced. One such approach is Modified Perturbed Sampling for LIME (MPS-LIME) \cite{Shi2020}, which addresses the limitations of standard sampling in LIME. MPS-LIME considers feature correlations by converting the superpixel image into an undirected graph and formulating perturbed sampling as a clique set construction problem. Experimental evaluations on Google's pre-trained Inception neural network \cite{7298594} demonstrate that MPS-LIME surpasses LIME regarding understandability, fidelity, and efficiency, offering significantly improved explanations for black-box models. While the current work shares similarities with \cite{Shi2020} in utilizing a graph representation, it incorporates additional considerations with a prominent focus on stability.

The proposed algorithm distinguishes itself from previous works by offering a purposeful sample generation approach that addresses stability, improves localization precision in bounding boxes, and considers feature correlations. It comprehensively improves on all these factors, significantly contributing to the field of perturbed-based XAI models. Unlike other approaches, it achieves simultaneous improvements in stability, explanation qualities, and run-time efficiency. Additionally, our method does not require additional training data and is simple to implement while enhancing LIME in multiple aspects. Notably, the stability enhancement provided by our method is guaranteed. Unlike solutions limited to tabular data, our approach is applicable to image data as well. Furthermore, we conducted experiments on a larger dataset, utilizing more suitable metrics such as Jensen-Shannon Divergence (JS\_DIV) \cite{10.5555/3365202} to compare bounding box similarities between ground truth data and explanations, even in cases without overlap.

\section{Methodology}\label{section:3}

In this section, we outline the random sampling algorithm of LIME, followed by a detailed overview of our MindfulLIME algorithm, which employs a purposive sample generation-based approach. Our enhanced perturbation-based XAI algorithm framework is crafted to function with any data that can be transformed into a graph of correlated features. Specifically, in this paper, we underscore its application and validation employing a dataset of X-ray images, wherein each segment of an image is treated as a feature and represented as a vertex in the graph.

\subsection{LIME}

LIME, which stands for Local Interpretable Model-agnostic Explanations, is an innovative method designed to explain the predictions of various types of machine learning models in an interpretable and faithful manner. Before diving into the LIME framework, it is essential to understand the initial processing of the input image. Image classification tasks typically involve working with tensors with three color channels per pixel. However, interpreting such tensors can be challenging and computationally intensive. This is where superpixel-based representations come into play. Images are composed of individual pixels, but it is important to note that a single pixel may not convey a complete and accurate representation of the depicted scene. Therefore, selecting groups of pixels that collectively represent semantically meaningful features becomes necessary, particularly for tasks like object detection. Segmentation algorithms are designed to group pixels based on common properties such as color, forming superpixels. Superpixels are non-overlapping segments that are easier to work with than individual pixels. By converting pixels into superpixels, we increase expressiveness while reducing complexity. In the context of the LIME framework, superpixels are treated as features \cite{pmlr-v139-garreau21a}.

\begin{figure*} %[ht]
    \centering
    \includegraphics[width=\textwidth]{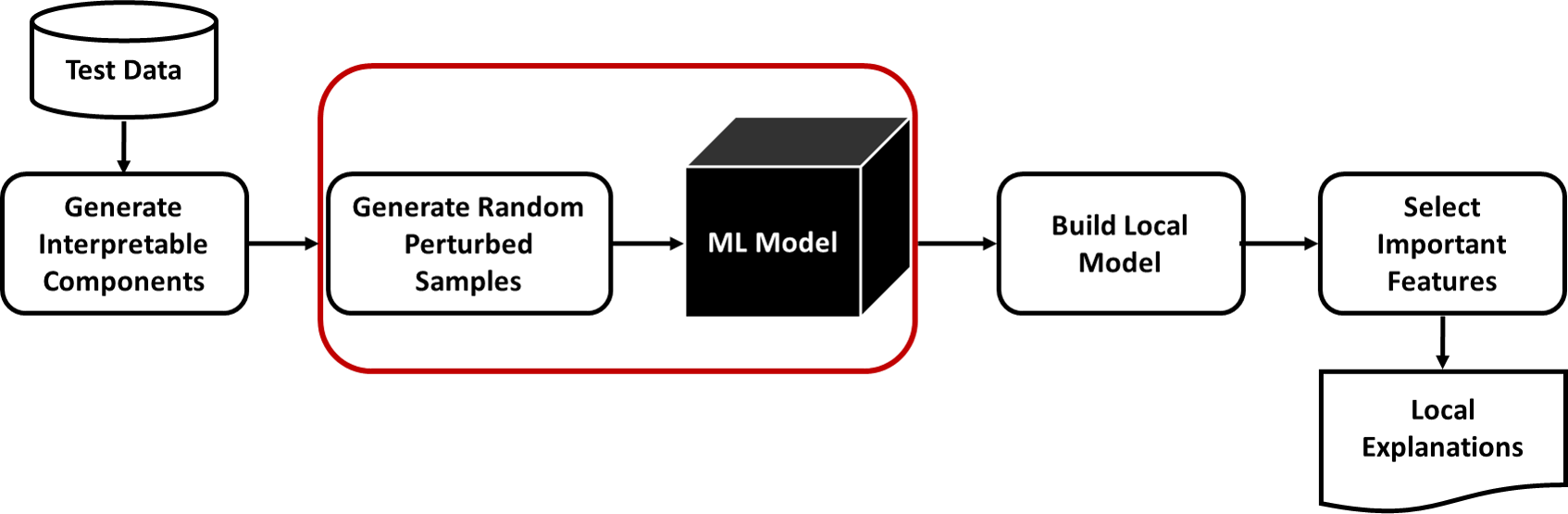}
    \caption{A block diagram illustrating the high-level structure of the LIME framework}\label{fig:1}
\end{figure*}

Figure \ref{fig:1} presents a high-level block diagram illustrating the LIME framework. The initial step involves selecting a specific instance for which we want to explain the classifier's decision. LIME then generates perturbed samples by introducing random disturbances to the features of the selected instance, as generated in the preceding step. Specifically, this involves turning on and off different superpixels. LIME aims to gain insights into the local behavior of a black box model $f$ around the point of interest $x$, even without knowing its internal structure.

Given the lack of detailed information about the inner workings of the black box model $f$, LIME tackles this challenge by randomly generating samples in the proximity of $x$ and querying the model to obtain predictions for each generated sample (red rectangle in Figure \ref{fig:1}). This process creates a simulated dataset $D = {(x_{1}, y_{1}), (x_{2}, y_{2}), ..., (x_{n}, y_{n})}$, where $y_{i}$ represents the class probabilities outputted by $f$ for $x_{i}$. The number of perturbations used in the sampling process is determined by the hyperparameter $n$. The model $f$ can take various forms, such as regression (with $y_{i} \in R$) or classification (with $y_{i} \in {0, 1}$ or $y_{i} \in [0, 1]$ for probability outputs). To provide an explanation, LIME chooses a linear model $g$ (denoted as the local model in Figure \ref{fig:1}) from a collection of function spaces $G$, by solving an optimization problem:

\begin{equation}
    \underset{g \in G}{\mathrm{argmin}} L(f,g,\pi _{x})+\Omega(g)
\end{equation}

\noindent Where $f$ refers to the original (or black box) model, $x$ represents the original data point on which the explanation will be performed, and $g$ corresponds to the model deduced by LIME, which can vary from a small linear model to a decision tree or a rule list, contingent on the nature of the data. This selection is based on the widely held belief that, while not universally interpretable across all data types, simpler models are often easier for users to understand.
The proximity measure $\pi$ quantifies the distance between an instance $x_i$ and $x$, defining the locality around $x$. It assigns weights to $x_i$ (perturbed instances) based on their distance from $x$. The first term of the formula shows the measure of the unfaithfulness of $g$ in approximating $f$ in the locality defined by $\pi$. This is known as the locality-aware loss in the original paper. The second term represents a measure of the complexity of the explanation model $g$. The coefficients of the linear model $g$ indicate the contribution of each superpixel towards the prediction of each class \cite{zhang2019why}.

To summarize, LIME introduces slight modifications to the features of the input data point and generates new samples. For each perturbed image, LIME queries the model being explained to receive the class probabilities and then trains a linear model to approximate the class probabilities of the perturbed images based on the presence or absence of each superpixel. The coefficients of this linear model indicate the contribution of each superpixel towards the prediction of each class. By systematically modifying the value of a variable, LIME computes the difference between the model's output for the new, modified data point and its output for the original input data point. This information is then used to determine the importance of that particular variable. In the context of this study, the variables that LIME selectively retains or disregards in each iteration correspond to the features of the generated samples, which are segments of image instances. For image data, the explanation provided by LIME typically involves highlighting the superpixels associated with the top positive and negative coefficients of the linear model, indicating which parts of the image have the greatest positive and negative influence on the model's prediction.\cite{pmlr-v139-garreau21a}

\subsection{MindfulLIME}

In this subsection, we present our purposive perturbed sampling algorithm, which primarily aims to address the instability of LIME output. As mentioned earlier, LIME generates samples randomly, posing risks such as output instability, generation of out-of-distribution data, or even creating nonsensical samples that disrupt the learning of local explanation models. In contrast, our algorithm adopts a deterministic approach, ensuring stability in the generated output. A deterministic algorithm follows a specific set of rules or instructions, consistently producing the same output for a given input. We control the sampling process by employing a non-random algorithm, allowing us to retain samples and consider specific conditions during their generation selectively. This capability significantly enhances the quality of the generated samples efficiently.

\begin{figure*}
    \centering
    \includegraphics[width=\textwidth]{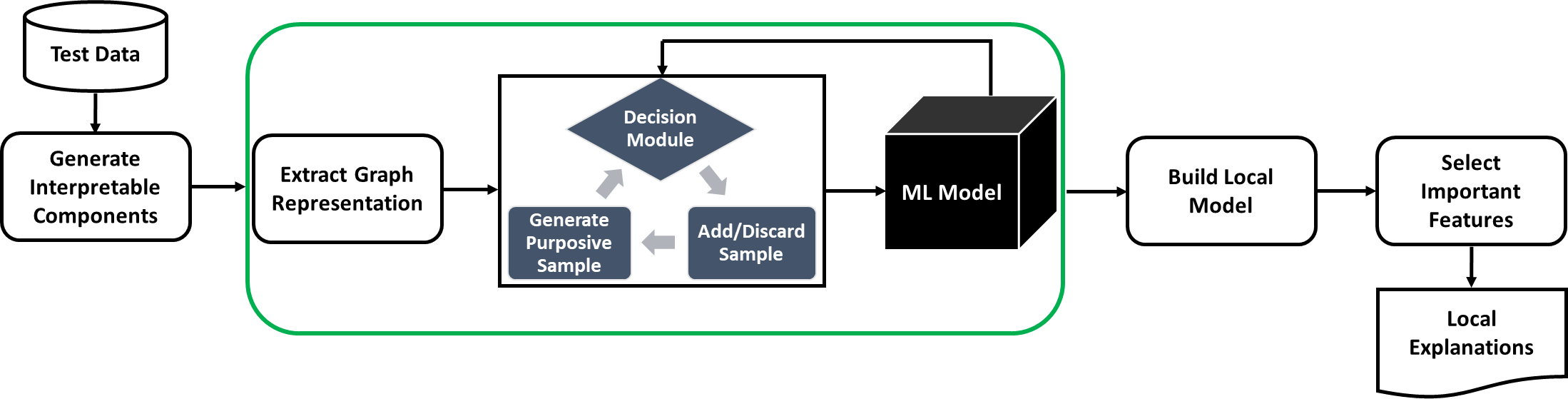}
    \caption{A block diagram illustrating the high-level structure of the MindfulLIME framework}\label{fig:2}
\end{figure*}

Figure \ref{fig:2} illustrates a high-level block diagram showcasing the MindfulLIME framework. 

In this framework, the instance that needs an explanation is represented as a graph, where the superpixel segments are depicted as vertices, and only the adjacent segments are connected by edges. Superpixels are generated using popular segmentation algorithms, such as Quickshift \cite{vedaldi2008quick}, Felzenszwalb \cite{felzenszwalb2004efficient}, and SLIC \cite{achanta2012slic}, the details and comparisons of which will be provided in \ref{Quant} Section.

We denote this graph as $G = (V, E)$, where $V$ represents the set of vertices (superpixels) and $E$ represents the set of undirected edges. The cardinalities of $V$ and $E$ are denoted as $\left | V \right | = S$ (indicating the number of superpixels) and $\left | E \right |$, respectively.

Our modified sampling solution is presented in the following algorithms. The core idea is to systematically analyze the graph of superpixels and generate new samples by selectively deactivating a specific subset of superpixels while leaving the rest activated. We represent the superpixel mask as a binary vector, denoted as $mask$, with a size of $S$, where each index corresponds to an individual superpixel, denoted as $s_i$. A value of $1$ indicates an activated superpixel, while $0$ represents a deactivated one. In this context, ``activate'' refers to keeping the superpixel in its original state, while ``deactivate'' involves substituting the pixel values with the mean of all pixels belonging to that particular superpixel.

There are two phases within Algorithm \ref{algo1}, \emph{PurposiveSampleGeneration}. In the initial phase, we generate masks where only one superpixel is deactivated per mask. This process aims to identify the individual superpixels that have a negative impact on the certainty of the classifier. Our objective is to determine which superpixels, in this phase, or a combination of superpixels in the next phase, when deactivated, cause the data point to deviate from the expected distribution. In the context of this study, when deactivating one or more superpixels decreases the certainty of the trained classifier, it undermines the reliability of the classifier's output for the corresponding generated sample. This reduction in certainty serves as an indicator of an Out-of-Distribution generated sample, the exact nature of which will be elaborated upon later in this section.

Assuming superpixels are represented as vertices, we generate $V$ masks in this step, where each mask has only one deactivated vertex with a value of $0$, while the rest of the vertices have a value of $1$. The new mask is then applied to the original image, and the modified image is passed to the \emph{DecisionModule}, Algorithm \ref{algo2}. The new mask is either discarded or retained depending on the decision the \emph{DecisionModule} returns. If the new mask is discarded, the deactivated vertices and their connected edges are removed from the original graph. When the new mask is retained, a new sample tuple is generated and appended to the \emph{Generated Sample Table (G)}, which includes mask values. Each \emph{sample} has three properties: \emph{path}, \emph{value}, and \emph{processed}. The \emph{path} property consists of a list of edges $(v_i, v_j)$ that represent the path and the order in which the deactivated vertices in the corresponding sample are processed. The \emph{value} property represents the mask value of the sample. The binary \emph{processed} property indicates whether the sample has been processed or not, based on whether the second vertex $v_j$ in the last appended edge $(v_i, v_j)$ in its \emph{path} has been visited or not.

A subgraph consisting of the retained vertices from the first phase is passed as input to the second phase. In the second phase of Algorithm \ref{algo1}, we iterate over each generated sample, which is appended to the \emph{Generated Sample Table ($G$)} (outer loop). If it has not been processed before, we retrieve the second vertex from the last appended edge in its \emph{path} property for each sample. We then examine the unvisited adjacent vertices of the current vertex. In each iteration of the nested loop, we select one of the adjacent vertices, denoted as $v_j$, and generate a new mask by modifying the previous mask. This modification involves deactivating the vertex $v_j$ in the new mask. Similar to the first phase, a new sample with the value of the new mask is generated and retained if determined to be appropriate by the \emph{DecisionModule}. Otherwise, it is ignored. This process continues until no new samples can be generated and all the generated samples are processed.

\begin{algorithm}[]
\caption{PurposiveSampleGeneration}\label{algo1}
\begin{algorithmic}[1]
\Require \textit{Undirected Edge List (E), Vertex List (V), original\_image, threshold} % , threshold
\Ensure \textit{Generated Sample Table (G)}
%\State $G \gets \left \{  \right \}$ 
\State $G \gets Table()$
\For{$v \gets 1$ \textbf{to} $V$}
    \State $mask \gets [1] * V.count$
    \State $mask[v] \gets 0$
    \State $sample\_image \gets apply\_mask(original\_image, mask)$
    \State $decision \gets DecisionModule(sample\_image, threshold)$
    \If {$decision == True$}
        \State $new\_sample \gets Sample()$ 
        \State $new\_sample.path.append((v,v))$ 
        \State $new\_sample.value, new\_sample.processed \gets mask, False$ 
        \State $G.append(new\_sample)$
    \Else
        \State $V.remove(v)$
        \For{$(x,y)$ \textbf{in} $E$}
            \If{($x == v$ \textbf{or} $y == v$)}
                \State $E.remove(x,y)$
            \EndIf
        \EndFor
    \EndIf
\EndFor
\For{$sample$ \textbf{in} $G.itertuples()$}
    \If{$sample.processed == False$} % \textbf{and} $sample.deleted == False$
        \State $v \gets second\_vertex(last\_edge(sample.path))$
        \For{$(x,y)$ \textbf{in} $E$}
            \If{($x == v$ \textbf{or} $y == v$)}               
                \If{($(x, y)$ \textbf{not} \textbf{in} $sample.path$) \textbf{and} ($(y, x)$ \textbf{not} \textbf{in} $sample.path$)}
                    \State $mask \gets sample.mask$
                    \State $mask[x], mask[y] \gets 0, 0$
                    \State $sample\_image \gets apply\_mask(original\_image, mask)$
                    \State $decision \gets DecisionModule(sample\_image, threshold)$
                    \If {$decision == True$}
                        \State $new\_sample \gets Sample()$ 
                        \State $new\_sample.path.append((x,y))$ 
                        \State $new\_sample.value, new\_sample.processed \gets mask, False$ 
                        \State $G.append(new\_sample)$
                    \EndIf
                \EndIf
            \EndIf
        \EndFor
        \State $sample.processed \gets True$
    \EndIf
\EndFor
\end{algorithmic}
\end{algorithm}

The \emph{DecisionModule}, Algorithm \ref{algo2}, plays the main role in determining whether the newly generated sample should be retained or discarded. Its primary objective is to identify and select In-Distribution data points, which are characterized by higher confidence from our trained classifier. We employ a technique known as uncertainty sampling to filter out potential Out-of-Distribution data points. Uncertainty sampling is a machine learning and data annotation approach that prioritizes samples for further analysis or labeling based on their level of uncertainty. It utilizes probabilistic models, such as classifiers or predictive models, to estimate the uncertainty or confidence associated with each sample. By actively seeking out uncertain samples, uncertainty sampling aims to enhance the model's performance by focusing on areas where additional labeled data or targeted analysis would likely yield the greatest benefits.

\begin{algorithm}[]
\caption{DecisionModule}\label{algo2}
\begin{algorithmic}[1]
\Require \textit{sample\_image, threshold}
\State $probability \gets Classifier(sample\_image)$
\If {$probability > threshold$}
\State \Return $True$
\EndIf
\State \Return $False$
\end{algorithmic}
\end{algorithm}

There are various methods of uncertainty sampling, including entropy-based sampling, margin sampling, and confidence-based sampling. Entropy-based sampling selects samples based on the entropy of the predicted class probabilities. Higher entropy indicates higher uncertainty, as the model is less confident in its predictions \cite{10.1007/978-1-4471-2099-5_1}. Margin sampling, on the other hand, considers the difference between the top two predicted probabilities. A smaller margin suggests higher uncertainty, as the model is less decisive between the top two classes \cite{10.1162/153244302760185243}. Lastly, confidence-based sampling is a straightforward approach that includes samples if their predicted probability for the selected class exceeds a predefined confidence threshold \cite{culotta2005reducing}. In this work, we applied this method in a reverse manner by focusing on selecting samples for which the model exhibits high confidence while disregarding those with lower probabilities \cite{Cardellino2015ReversingUS}.

Given that our problem involves multi-class, multi-label classification, each sample can belong to multiple classes, with each class potentially having multiple annotations. As a result, we execute the LIME algorithm separately for each class among the top three predicted probabilities to which the sample belongs. In Algorithm \ref{algo2}, we input the sample image into the trained classifier to calculate the probability of it belonging to the desired class. We need a criterion to determine which samples the model is confident about and consider as in-distribution data points. Confidence-based sampling is the most suitable uncertainty sampling technique for the current problem considering the limited information available.  

To select thresholds, we calculate the average probabilities for each class over a small random unseen subset of the CheXpert dataset, specifically when the classifier correctly detects the correct label. These average probabilities serve as thresholds in our sample generation algorithm for each class within the top three classes to which the original image belongs. Therefore, if the probability of a newly generated unlabeled sample belonging to the desired class exceeds the respective threshold, we consider it to have a lower likelihood of being an out-of-distribution data point, making it more suitable among some other samples.

This approach ensures that the order in which superpixel combinations are tested does not affect the stability of the outcome or efficiency of the algorithm. Moreover, it is important to note that while our algorithm does generate and evaluate multiple combinations of superpixels, it does so in an efficient manner. In the initial phase, the algorithm quickly narrows down the set of relevant superpixels by evaluating individual superpixels, and in the second phase, the algorithm intelligently generates and tests combinations of adjacent superpixels, rather than exhaustively testing all possible combinations. 

Our method not only reduces the number of samples that need to be generated and evaluated but also leverages the spatial relationships between superpixels to generate more informative explanations. In comparison to LIME, which requires generating a much larger number of samples to accurately estimate the local decision boundary, our algorithm terminates with fewer samples while achieving better or comparable results. Our experiments indicate that our algorithm outperforms LIME in terms of efficiency, while consistently delivering stable results. Detailed results of these experiments will be provided further in Section \ref{section:5}.

The time complexity of the LIME random perturbation technique is influenced by various factors, including the complexity of the underlying model being explained, the number of perturbed samples, and the number of features or dimensions in the data. In this study, the features correspond to superpixels. The time complexity of the LIME random perturbation technique can be approximated as $O(d \times n \times f)$, where $d$ represents the number of features (superpixels), $n$ represents the number of perturbed samples, and $f$ represents the complexity of fitting the interpretable model. This approximation assumes that the complexity of LIME is directly proportional to the number of perturbed samples $n$. Considering that the number of features $d$ is generally much lower than the number of samples $n$ ($|d|<<|n|$), we can simplify the complexity of LIME as $O(n)$, where $n$ represents the number of samples.

The complexity of MindfulLIME depends on the number of features (superpixels), represented as $d$, and the depth of the graph exploration. Initially, $d$ samples are created, each containing a single $0$ for the corresponding superpixel, resulting in an initialization complexity of $O(d)$. In the subsequent phase, individual adjacent superpixels to the previously visited ones are considered during each iteration. In the worst-case scenario, for every $d$ previously generated samples, $k$ new samples are generated. As a result, the complexity of MindfulLIME ($i$ levels) can be estimated as $O(dk^i)$. The values of $d$ and $k$ can vary depending on the selected superpixel algorithms, highlighting the need to investigate further their impact on the performance and efficiency of MindfulLIME. This investigation will be conducted in the following section.

\section{Experimental Setup}\label{section:4}

Our research focuses on generating explanations for the results of the thorax disease classifier within the domain of chest X-ray (CXR) images. Careful dataset preparation and classifier training are crucial initial steps. Additionally, having suitable metrics to assess the quality of explanation models is of utmost importance. The following subsections will present detailed information regarding the datasets, models, and metrics utilized.

\subsection{Datasets}\label{subsection:41}

This study uses two datasets: CheXpert \cite{article} and VinDr-CXR \cite{Nguyen2020}. These datasets were thoughtfully selected as the most suitable and complementary combination. CheXpert offers sufficient samples per class and reliable global annotations, making it well-suited for training and classification using deep neural networks. Conversely, vinDr-CXR provides precise regional annotations for classes that align with CheXpert, making it a valuable resource for evaluating explanations. CheXpert ensures adequate training samples, while vinDr-CXR provides essential and precise local annotations for evaluation.

\begin{figure*} %[ht]
\centering
\begin{tabular}{c c}
  \includegraphics[width=0.40\textwidth]{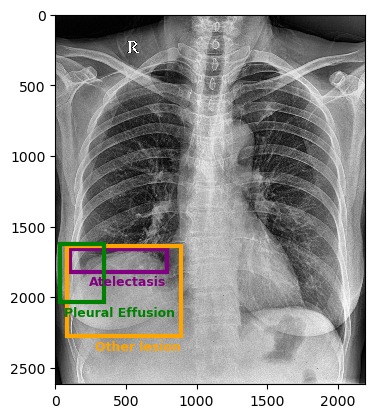} &
  \includegraphics[width=0.38\textwidth]{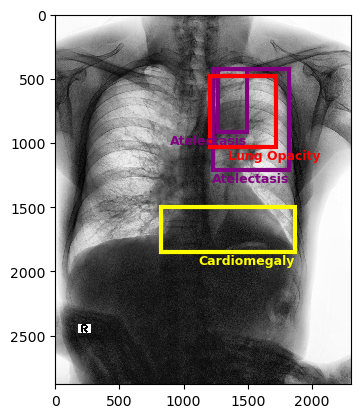}
\end{tabular}
\caption{Samples of the VinDr-CXR dataset with precise annotated labels}
\label{fig:3}       
\end{figure*}

The CheXpert dataset comprises 224,316 chest radiographs from 65,240 patients, accompanied by chest-related global annotations. These annotations were extracted using an automated, rule-based labeler that parsed categories from the free-text radiology reports authored by radiologists. To ensure reliability, additional rules were applied to capture negations and uncertainties mentioned in the reports. Each sample within the CheXpert dataset includes the image ID, patient's sex and age, X-ray properties (frontal/lateral and AP/PA), and a hot vector representing 14 distinct classes. Within this vector, 0 labels indicate negative findings, +1 labels represent positive findings, and -1 labels indicate uncertain findings \cite{article}. 

The VinDr-CXR dataset includes 15,000 images within its training set, with labels manually defined by a team of radiologists. Figure \ref{fig:3} presents sample images from this dataset. The dataset includes 22 local labels represented by bounding boxes surrounding abnormalities, and six global labels denoting detected diseases. The images are provided in the Digital Imaging and Communications in Medicine (DICOM) format, accompanied by relevant DICOM tags such as patient sex and age. Primarily consisting of posterior-anterior (PA)-view chest X-rays, the dataset excludes non-relative X-rays from other body parts \cite{Nguyen2020}. 

\subsection{Training The Classifier}\label{subsection:42}

The accuracy of the classifier's results greatly influences the explanations' quality. Therefore, our trained model must identify labels accurately. Explanations generated by a poorly performing classifier are often uninformative and lack value. We apply data preprocessing techniques to the CheXpert dataset to ensure data quality. We exclude samples that contain uncertain labels among their classes. Furthermore, we only select PA and Lateral images for both sets. The test set is not yet available because the CheXpert dataset is an ongoing competition. Hence, we conduct the training phase using 85\% of the samples from the filtered training set (83,781) while reserving the remaining 15\% for validation and testing purposes. Within the remaining 15\%, 10\% is allocated for validation, and the remaining 5\% is combined with the filtered original validation test to form the test set. 

Subsequently, we train several well-known CNN models, including InceptionV3 \cite{szegedy2016rethinking}, DenseNet121 \cite{huang2017densely}, VGG16 \cite{DBLP:journals/corr/SimonyanZ14a}, ResNet50 \cite{he2016deep}, MobileNetV2 \cite{sandler2018mobilenetv2}, and InceptionResNetV2 \cite{langkvist2014review}. All models are trained on an Nvidia GeForce RTX 3080 GPU, employing a batch size of 16. We utilize the Adam optimizer with a learning rate 0.0001, initializing the models with weights pre-trained on the ImageNet dataset. In order to determine the most suitable classifier model, we assessed model accuracy using the Area under the ROC Curve (AUC). Considering the overall performance, InceptionV3 was an excellent choice among all by AUC values for the selected classes as; Cardiomegaly 82\%, Pleural Effusion 92\%, Lung Opacity 94\%, Consolidation 91\%, Atelectasis 77\%, Pneumothorax 95\%, No Finding 92\%. This results in a mean AUC of 89\%. Additionally, the mean accuracy of the model is 88\%. As a result, we opted to continue the study with InceptionV3 as our selected classifier model.

\subsection{Evaluation Metrics}\label{subsection:43}

In this subsection, we explain the measurements we used to assess how well our recommended algorithm performs in terms of stability and accuracy. We convert the ground truth superpixel masks and explanation superpixel masks into one-dimensional lists to facilitate the analysis. This conversion enables us to calculate pixel-wise measurements and simplifies the handling of multi-bounding boxes for both types of masks. Moreover, we convert the pixel values into binary values to ensure comparability and consistency. Let us denote the ground truth mask as $GT$, the obtained explanation map as $EX$, and the total number of pixels as $P$. The evaluation metrics employed in this study are as follows:\\

\noindent\textbf{Stability:} To assess the stability of the explanations, we utilize the Intersection over Union (IOU) or Jaccard Index, a widely used metric in object detection tasks \cite{jaccard1912distribution, everingham2010pascal}. The IOU compares two masks, denoted as $GT$ and $EX$, to quantify the degree of overlap between them. It is calculated by dividing the estimated area's and ground truth areas' intersection by their union. A high IOU value of $1$ indicates a strong similarity between $GT$ and $EX$, signifying complete overlap between bounding boxes. Conversely, an IOU value of $0$ occurs when the intersection $|GT \cap EX|$ is $0$, indicating a lack of overlap between $GT$ and $EX$. The current work calculates the stability score by averaging the IOU values between the generated explanations and the actual bounding boxes over ten runs. The IOU can be computed using the following formula:

\begin{equation}
    IOU(GT,EX) = \frac{\sum_{i=1}^{P}GT(i)\cap EX(i)}{\sum_{i=1}^{P}GT(i)\cup  EX(i)}
\end{equation}

\noindent
\\\textbf{Localization Precision:} By comparing the localization of explanations to the ground truth annotations of the \textit{VinDr-CXR} dataset, we can assess different explanation methods based on their accuracy. While similarity metrics such as IOU, MAE \cite{mohseni2021quantitative}, and PWP (IoSR) \cite{li2021experimental} are commonly used, they often fail to capture key aspects like distance and proximity, especially when there is no overlap between the explanation and ground truth bounding boxes. In a recent study \cite{rahimiaghdam2024evaluating}, JS\_DIV was introduced as a more effective metric for evaluating similarity between distributions, particularly in scenarios with minimal or no overlap. Through comprehensive analysis based on criteria such as informativeness, coverage, localization, multi-target capturing, and proportionality, JS\_DIV consistently outperformed other metrics, offering a more balanced and precise measurement of divergence. This makes JS\_DIV particularly suitable for ensuring higher-quality visual explanations in medical imaging, where traditional metrics like IOU and MAE often fall short.

Jensen-Shannon Divergence (JS\_DIV) is a metric that quantifies the difference or similarity between two probability distributions \cite{lin1991divergence}. It is based on the Kullback-Leibler Divergence (KL\_DIV), which measures the extent to which one probability distribution deviates from another \cite{10.5555/3365202, bylinskii2018different}. In our case, we convert the $GT$ (ground-truth) and $EX$ (explanation) maps into probability distributions, ensuring that the sum of all values in each map equals one. Subsequently, the KL\_DIV is calculated as follows:

\begin{equation}
    KL\_DIV(EX_{pd},GT_{pd}) = \sum_{i=1}^{P}EX_{pd}(i)  log(\frac{EX_{pd}(i)+\varepsilon }{GT_{pd}(i)+\varepsilon})
\end{equation}

The KL\_DIV score is not symmetric or linear and ranges from zero to infinity. A lower KL\_DIV score between $EX_{pd}$ and $GT_{pd}$ indicates that $EX_{pd}$ provides a better approximation of the ground truth $GT_{pd}$. Essentially, the score reflects the difference in probabilities assigned to events in $EX_{pd}$ compared to those in $GT_{pd}$. When the probability of an event in $EX_{pd}$ is high while the corresponding probability in $GT_{pd}$ is low, there is a significant divergence. 
Conversely, if the probability of an event in $EX_{pd}$ is low while the probability in $GT_{pd}$ is high, the divergence is still significant but less pronounced than in the first case. The JS\_DIV utilizes the KL\_DIV to calculate a normalized and symmetrical metric. The calculation of the JS\_DIV is as follows:

\begin{equation}
    JS\_DIV(EX_{pd},GT_{pd})= \frac{KL(EX_{pd},M)+KL(GT_{pd},M)}{2}
\end{equation}

\noindent
where $M$ represents the average of two distributions. In our analysis, we employ the complementary metric $1-JS_DIV$ as a measure of similarity between two distributions in the context of localization precision. This metric, called the localization precision score, provides scores ranging from zero to one using the base-2 logarithm. A score of one indicates identical distributions, while a score of zero represents maximally different distributions \cite{10.5555/3365202, bylinskii2018different, lin1991divergence}. 

\section{Results and Discussion}\label{section:5}

This section presents the results of the proposed MindfulLIME algorithm through two subsections: qualitative and quantitative analysis. In the qualitative analysis, we visually explore the algorithm's performance using two random samples, offering valuable insights into the effectiveness of MindfulLIME. The quantitative analysis compares the performance of LIME and MindfulLIME, focusing on stability, localization precision, and run-time efficiency. Various settings, including segmentation algorithms, different sample numbers, and features, comprehensively assess the algorithms' performance in diverse contexts.

\begin{figure*}
\centering
\begin{tabular}{c c}
  \includegraphics[width=0.40\linewidth]{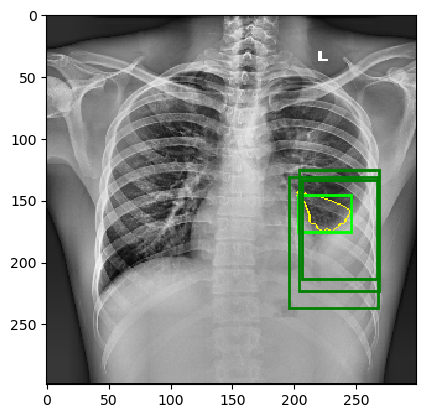} &
  \includegraphics[width=0.40\linewidth]{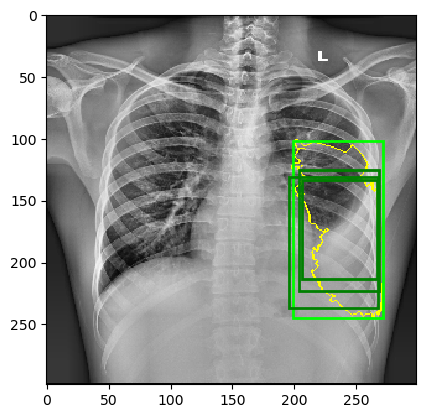}\\
  (a) MindfulLIME (1 Feature) & (b) MindfulLIME (4 Features) \\
  \includegraphics[width=0.40\linewidth]{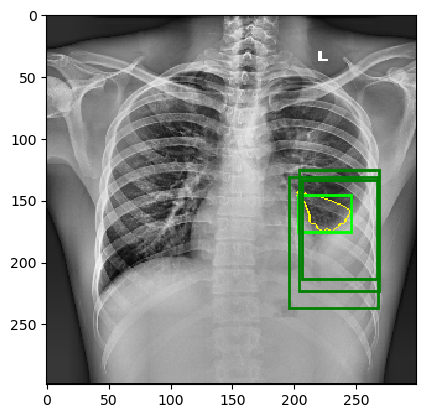} &
  \includegraphics[width=0.40\linewidth]{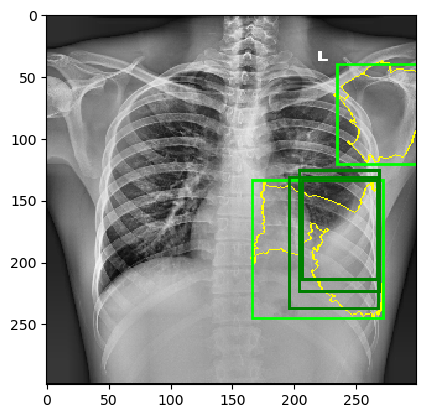}\\
  LIME 1000 Samples (1 Feature) & LIME 1000 Samples (4 Features)\\
  \includegraphics[width=0.40\linewidth]{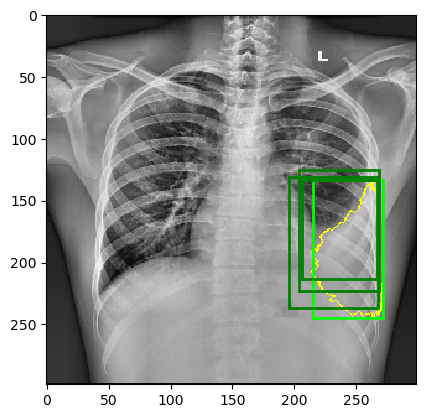} &
  \includegraphics[width=0.40\linewidth]{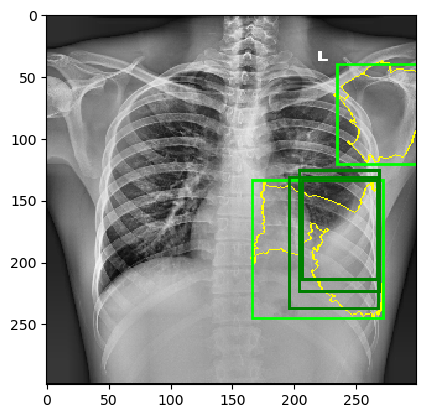}\\
  LIME 4000 Samples (1 Feature) & LIME 4000 Samples (4 Features)
\end{tabular}
\caption{Visual Analysis: Comparing generated explanations (the yellow-outlined areas enclosed with light green bounding boxes) by MindfulLIME and LIME to the ground truth annotations (dark green bounding boxes) for Random Sample 1}
\label{fig:4}       
\end{figure*}

\begin{figure*}
\centering
\begin{tabular}{c c}
  \includegraphics[width=0.40\linewidth]{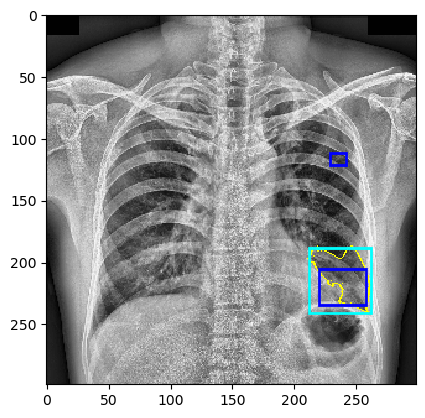} &
  \includegraphics[width=0.40\linewidth]{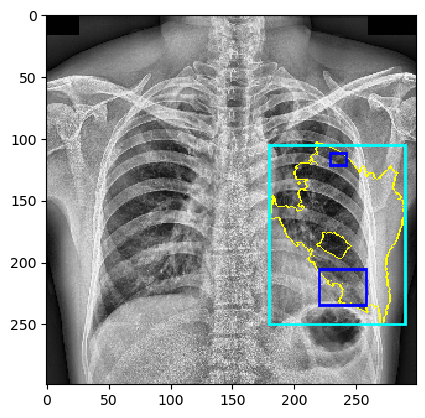}\\
  (a) MindfulLIME (1 Feature) & (b) MindfulLIME (4 Features) \\
  \includegraphics[width=0.40\linewidth]{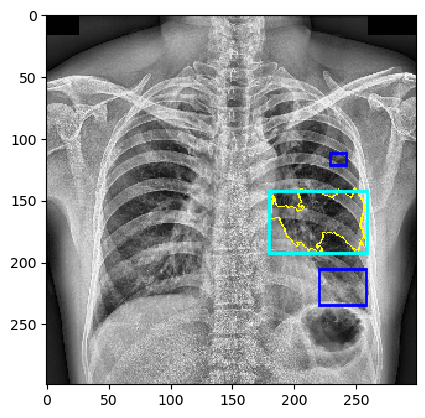} &
  \includegraphics[width=0.40\linewidth]{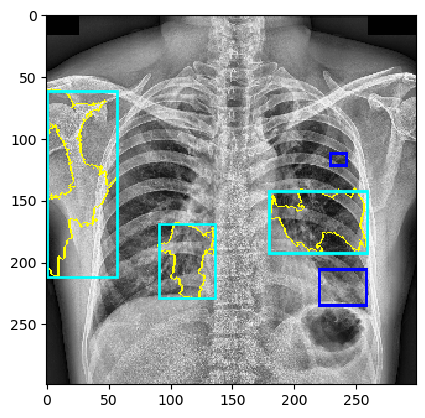}\\
  LIME 1000 Samples (1 Feature) & LIME 1000 Samples (4 Features)\\
  \includegraphics[width=0.40\linewidth]{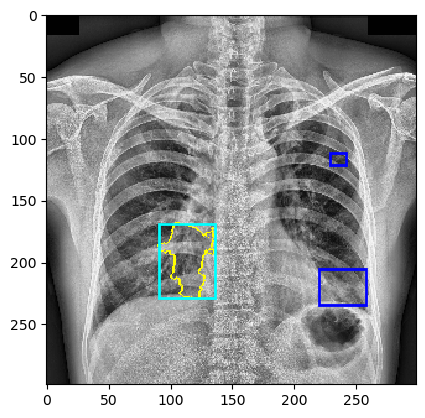} &
  \includegraphics[width=0.40\linewidth]{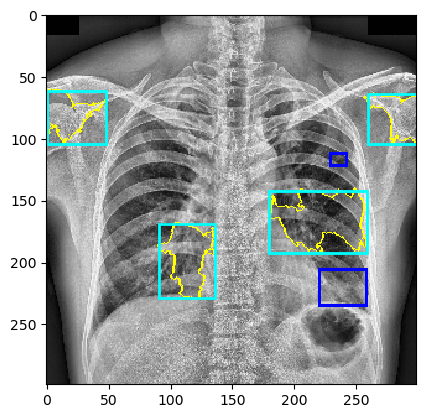}\\
  LIME 4000 Samples (1 Feature) & LIME 4000 Samples (4 Features)
\end{tabular}
\caption{Visual Analysis: Comparing generated explanations (the yellow-outlined areas enclosed with light blue bounding boxes) by MindfulLIME and LIME to the ground truth annotations (dark blue bounding boxes) for Random Sample 2}
\label{fig:5}       
\end{figure*}

\subsection{Qualitative Analysis}

The behavior of MindfulLIME compared to LIME is demonstrated in Figures \ref{fig:4} and \ref{fig:5}. Figure \ref{fig:4} showcases random sample 1 with a detected Pleural Effusion condition (annotations highlighted in green), while Figure \ref{fig:5} features random sample 2 with a detected Lung Opacity condition (annotations highlighted in blue). Each figure illustrates the disparities between the ground-truth annotations (darker-colored rectangles) and the generated explanations (lighter-colored rectangles surrounding the yellow border of the selected superpixels). We would like to emphasize that our trained classifier model generalizes well and is not simply memorizing the training data, as evidenced by its performance on a separate evaluation dataset.

The left columns depict the results for the top 1 feature, while the right columns display the results for the top 4 features. The first row presents the output of MindfulLIME, while the second and third rows represent the results of LIME using 1000 and 4000 samples, respectively. When comparing the results of LIME and MindfulLIME for two visual random samples, it becomes evident that MindfulLIME excels in multiple aspects. Firstly, MindfulLIME consistently identifies the top 1 feature with the highest similarity with the actual bounding box. Additionally, as the number of top features increases, MindfulLIME effectively prioritizes and presents more relevant features at the forefront. These findings highlight the enhanced performance and prioritization capabilities of MindfulLIME compared to LIME.

\subsection{Quantitative Analysis}\label{Quant}

We conduct experiments using a set of 200 test instances, running each for ten iterations to demonstrate the instability inherent in explanations generated by LIME. The experiments aim to explain predictions made by our trained chest X-ray classifier using both LIME and MindfulLIME. Our evaluation focuses not only on stability but also on precision and run-time efficiency, enabling a comprehensive comparison of the results.

The experiments involve evaluating three versions of MindfulLIME: MindfulLIME (2 levels), MindfulLIME (3 levels), and MindfulLIME (4 levels). Through experiments across different level settings, we examine the trade-off between performance and efficiency, gaining insights into the impact of levels on the algorithm's effectiveness and efficiency. We maintain the default values for most LIME parameters and settings, except for the generated random sample number and the segmentation algorithms used. We specifically focus on reporting the results of LIME for the most significant 1 and 4 features. The feature count, representing the number of patches (superpixels) generated by the LIME algorithm, necessitates manual configuration.

The default value for the number of perturbed samples in LIME is 1000. However, we also conduct experiments with 2000 and 4000 samples to investigate their impact on stability and other aspects. Additionally, we explore the effects of various popular segmentation algorithms apart from LIME's default algorithm (Quickshift). Each superpixel algorithm exhibits variations in shape, size, and number of generated segments. We aim to evaluate the influence of different superpixel shapes and numbers on precision and run-time efficiency.

\begin{figure*}
\centering
\begin{tabular}{c c c}
  \includegraphics[width=0.30\linewidth]{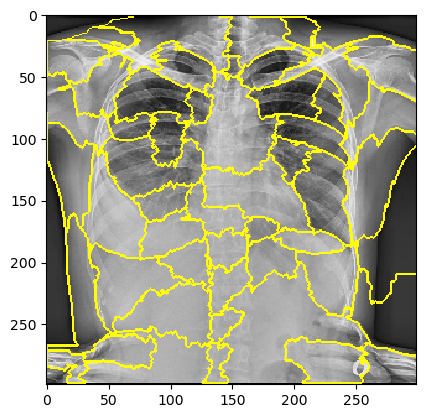} &
  \includegraphics[width=0.30\linewidth]{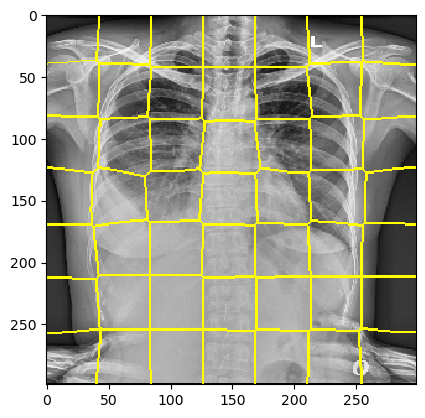} &
  \includegraphics[width=0.30\linewidth]{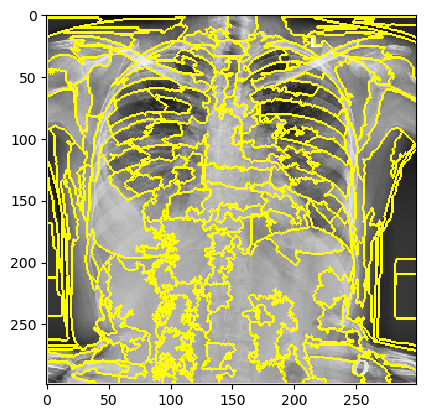}\\
  (a) Quickshift & (b) SLIC & (c) Felzenszwalb
\end{tabular}
\caption{Exploring Various Segmentation Algorithms: Given superpixels to the LIME generated by different segmentation algorithms}
\label{fig:6}       
\end{figure*}

We compare Quickshift \cite{vedaldi2008quick} with two additional segmentation algorithms: Felzenszwalb \cite{felzenszwalb2004efficient} and SLIC \cite{achanta2012slic}. Felzenszwalb is an efficient graph-based algorithm that utilizes the minimum spanning tree. SLIC, on the other hand, is a cluster-based superpixel algorithm that applies k-means clustering in a five-dimensional space of color information. Quickshift, the default algorithm in LIME, generates superpixels using an approximation of kernelized mean-shift through the mode-seeking method. To visually demonstrate these algorithms, Figure \ref{fig:6} shows the output of each algorithm on a random sample from the VinDr-CXR dataset.

We tune the parameters of the two additional segmentation algorithms to optimize their performance. We identify the most appropriate parameter values by conducting a comprehensive grid search on the parameter space of each segmentation method, using 50 randomly selected VinDr-CXR data samples. This thorough parameter tuning process enables us to tailor the segmentation algorithms to our specific data, ensuring they provide detailed and accurate results on X-ray images. The optimized parameter configurations for each segmentation approach are presented in Table \ref{table:1}.

\begin{table}
\centering
\caption{Segmentation Algorithms Parameters}\label{table:1}
\begin{tabular}{@{}lll@{}}
\toprule
Segmentation Algorithms & Parameters & Value \\
\midrule
SLIC & n\_segments & 50 \\
 & compactness & 80 \\
 & sigma & 20 \\
Felzenszwalb & scale & 600 \\
 & min\_size & 200 \\
 & sigma & 0.2 \\

\end{tabular}
\end{table}

We also determine the thresholds for sample generation by calculating the average probabilities for each class over a small random subset of the CheXpert dataset. Specifically, we focus on instances where the classifier accurately detects the labels of interest. The selected classes and their corresponding thresholds for the experiments are as follows: Cardiomegaly (0.42), Pleural Effusion (0.44), Lung Opacity (0.19), Consolidation (0.25), Atelectasis (0.16), and Pneumothorax (0.79). These thresholds serve as the criteria for identifying confident in-distribution data points during the sample generation process (Decision Module).

\begin{figure*}
\centering
\begin{tabular}{c}
  \includegraphics[width=0.45\linewidth]{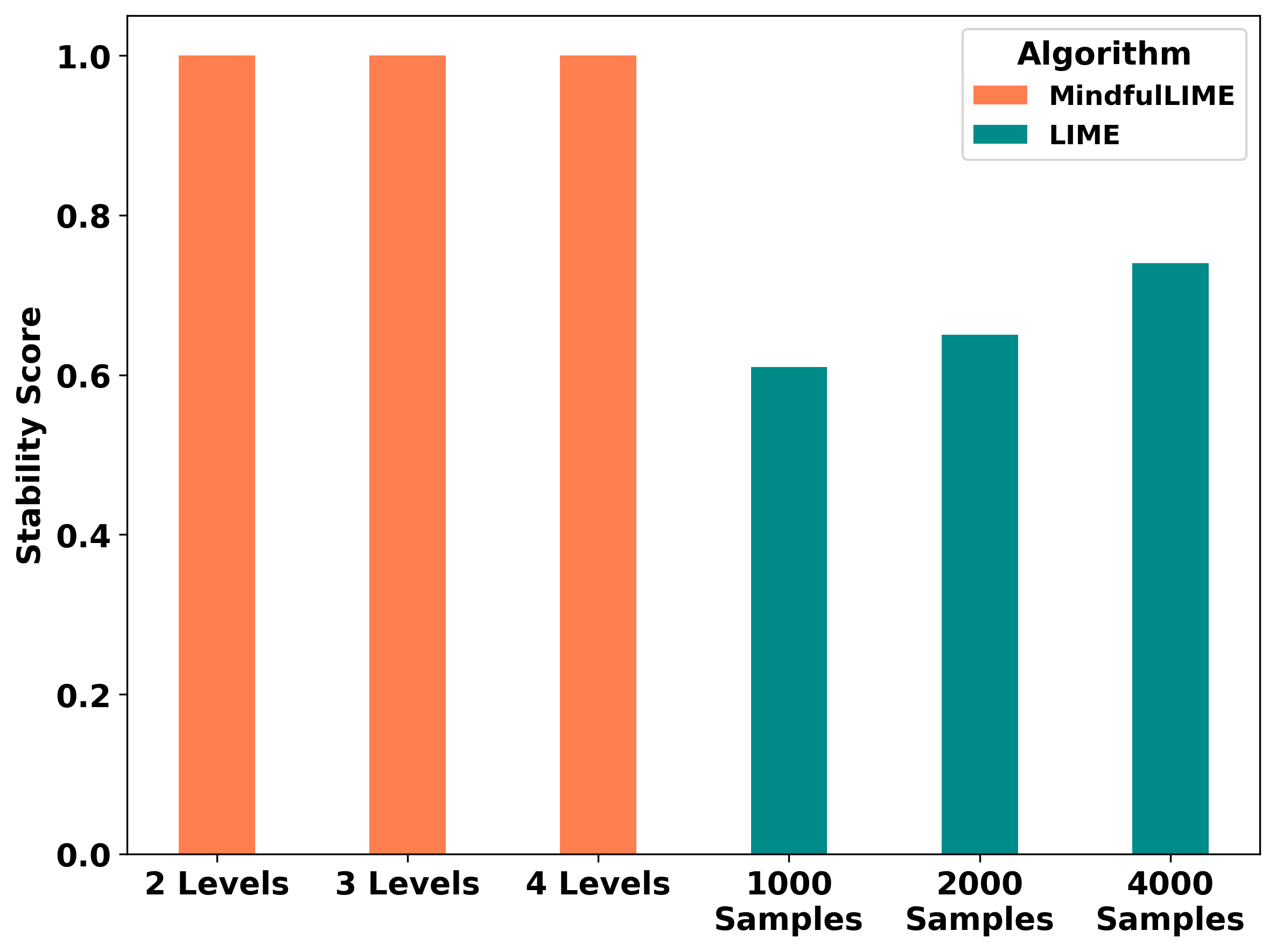}  \\
  (a) Quickshift \\
  \includegraphics[width=0.45\linewidth]{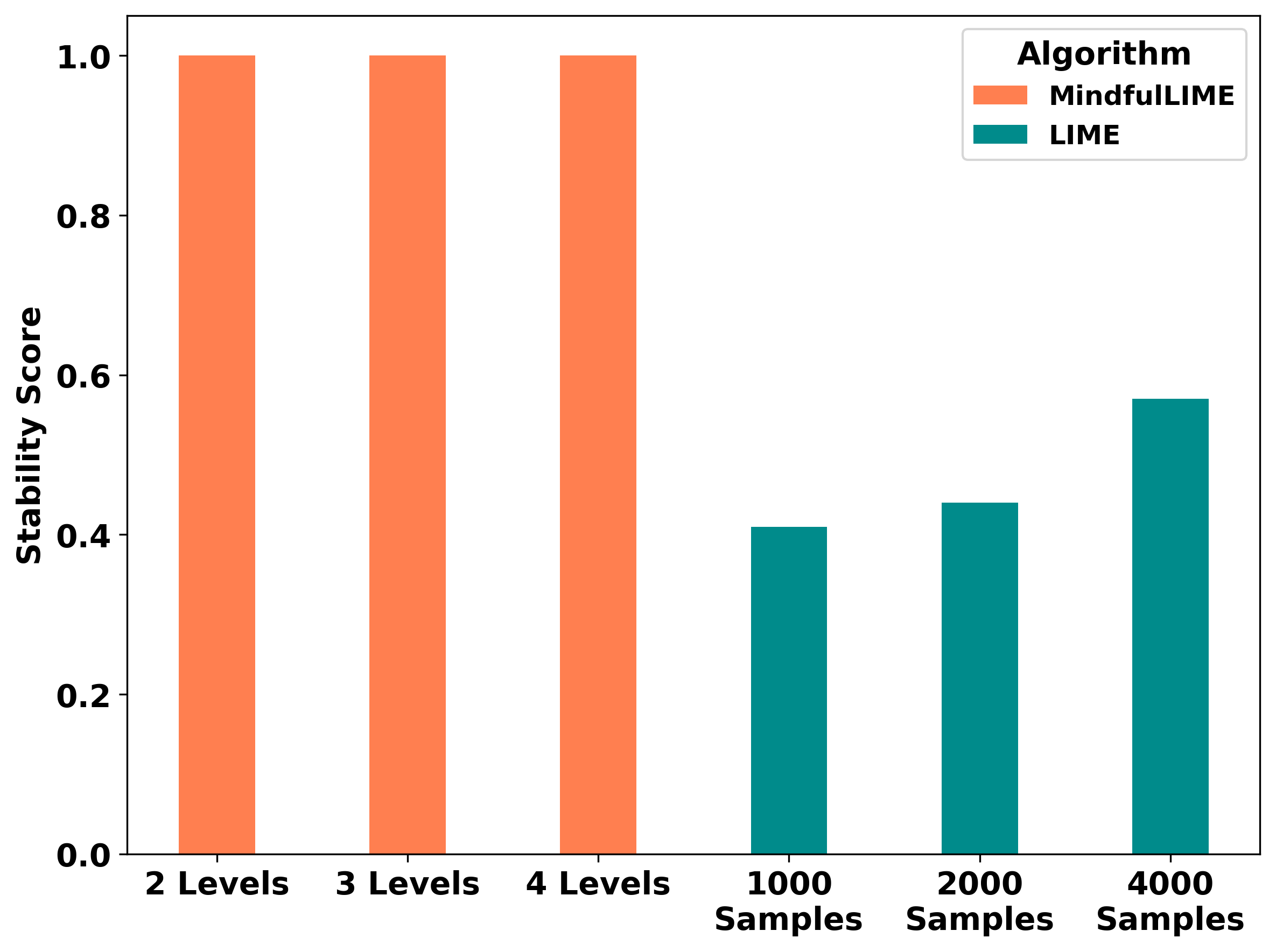}  \\
  (b) SLIC \\
  \includegraphics[width=0.45\linewidth]{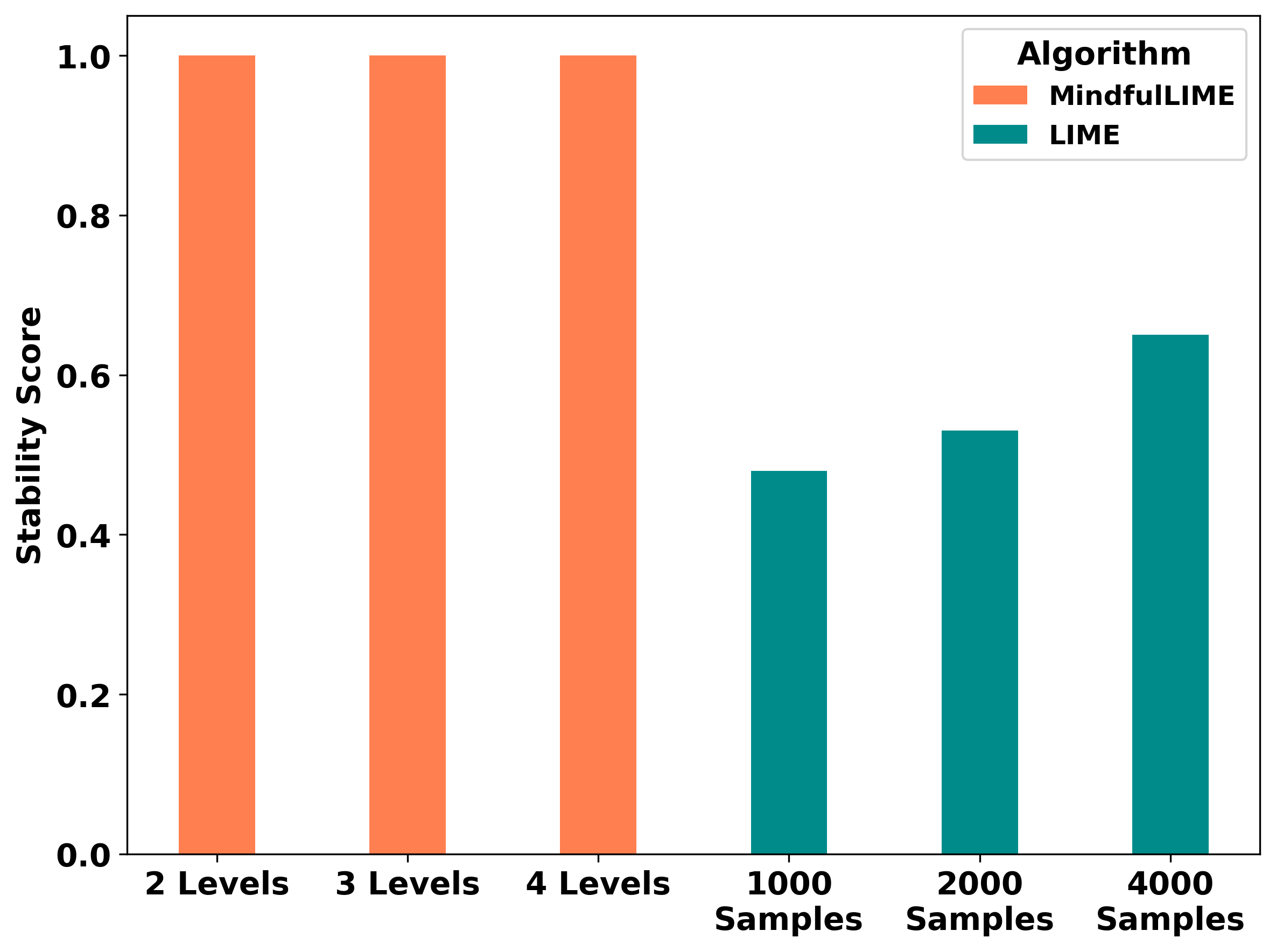}\\
  (c) Felzenszwalb \\
\end{tabular}
\caption{Stability Comparison: LIME vs. MindfulLIME for various versions and settings}
\label{fig:7}       
\end{figure*}

Figure \ref{fig:7} illustrates the stability results obtained for different versions of LIME and MindfulLIME using various superpixel algorithms. For each sample instance, we conducted the explanation generation task 10 times. We calculated the Intersection over Union (IOU) in each iteration between the ground truth and the generated explanation. The average IOU across all iterations was then computed as a quantitative measure of stability. Notably, only MindfulLIME achieved a perfect stability score of 100\%, while all other versions of LIME exhibited instability. Although increasing the number of random samples generated in LIME improved stability to some extent, the overall instability problem persisted. Regardless of the segmentation algorithm employed, MindfulLIME consistently demonstrated complete stability, confirming its expected effectiveness.

\begin{figure*}
\centering
\begin{tabular}{c}
  \includegraphics[width=0.45\linewidth]{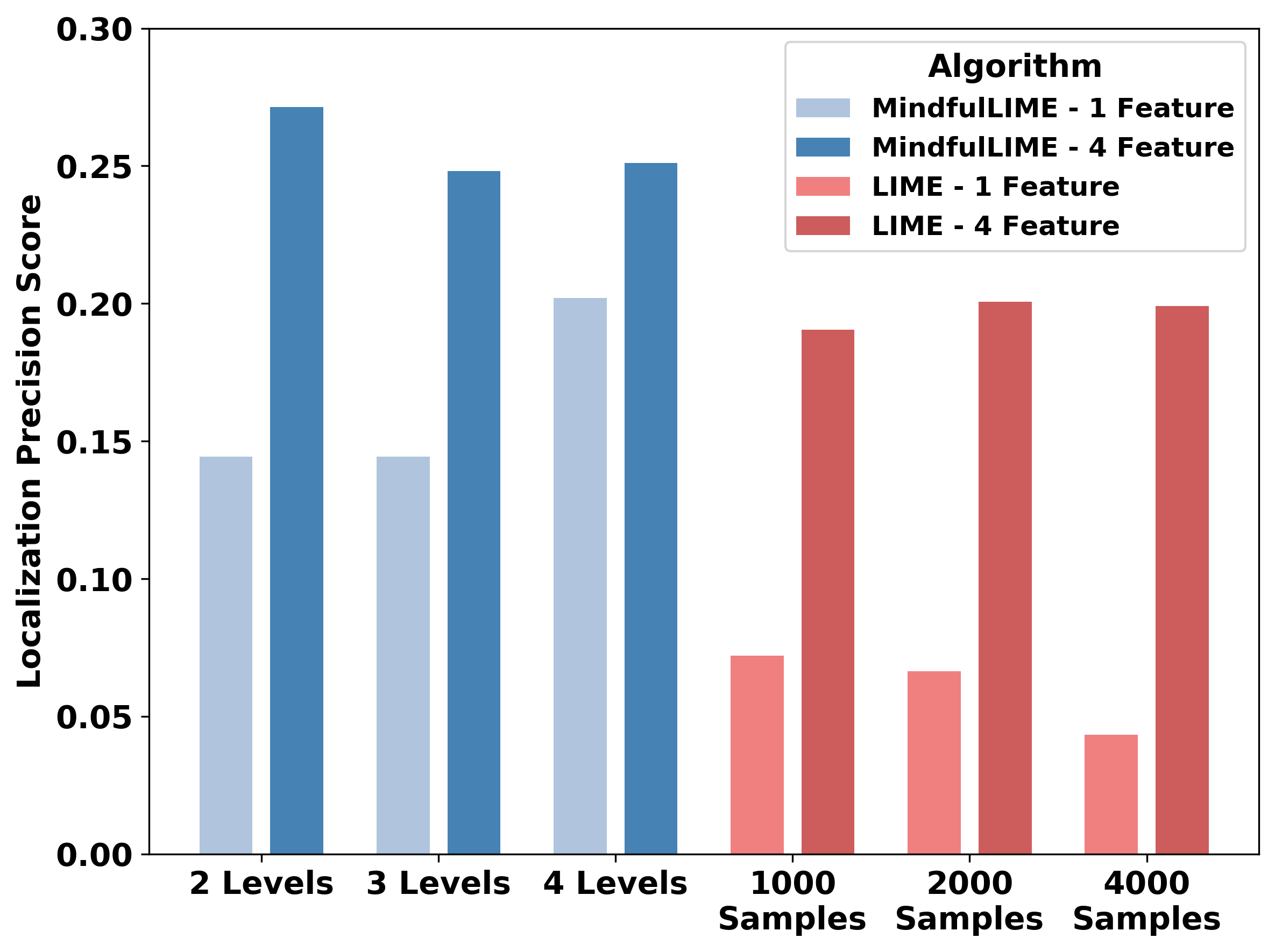} \\
  (a) Quickshift \\
  \includegraphics[width=0.45\linewidth]{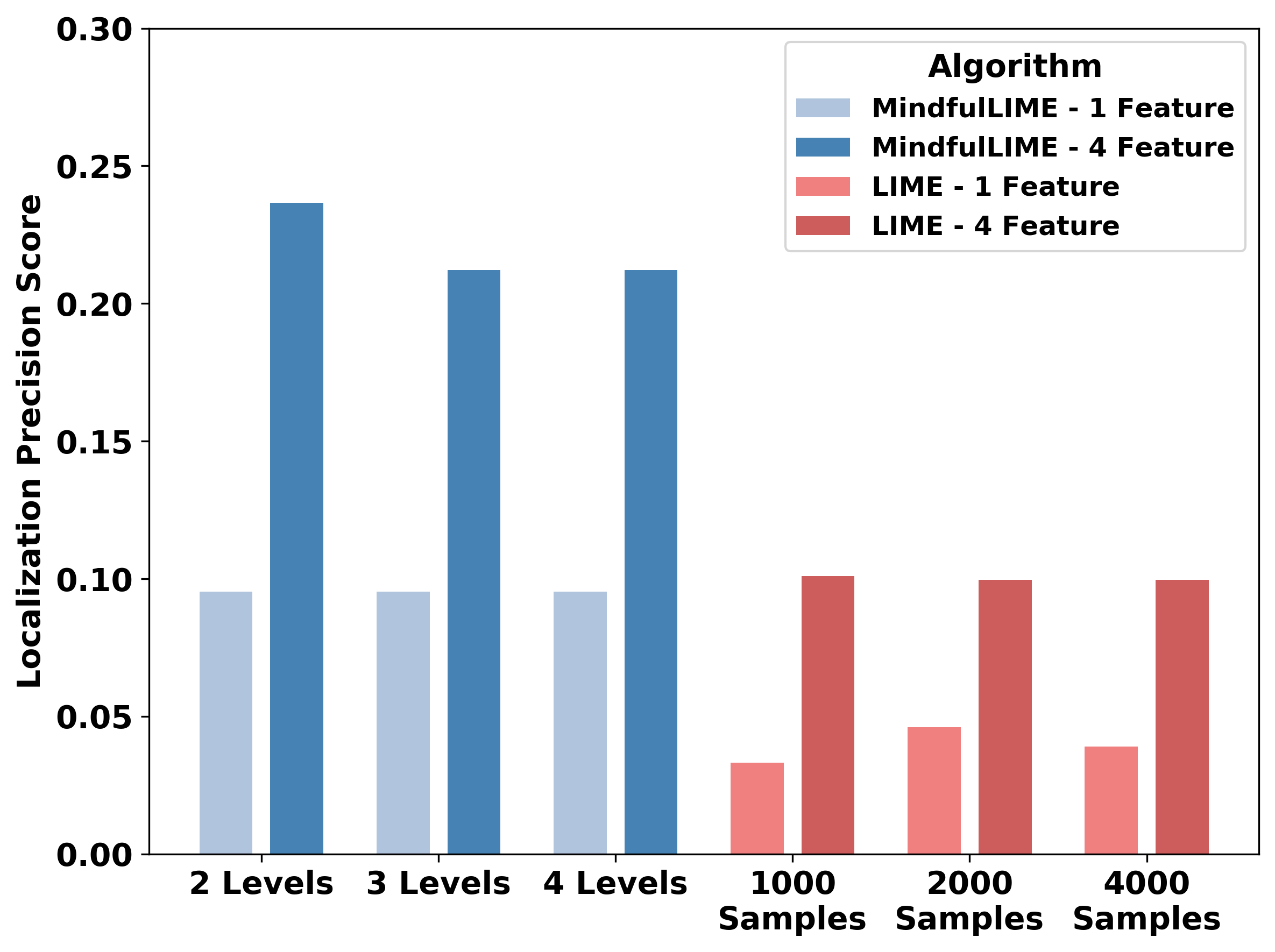} \\
  (b) SLIC \\
  \includegraphics[width=0.45\linewidth]{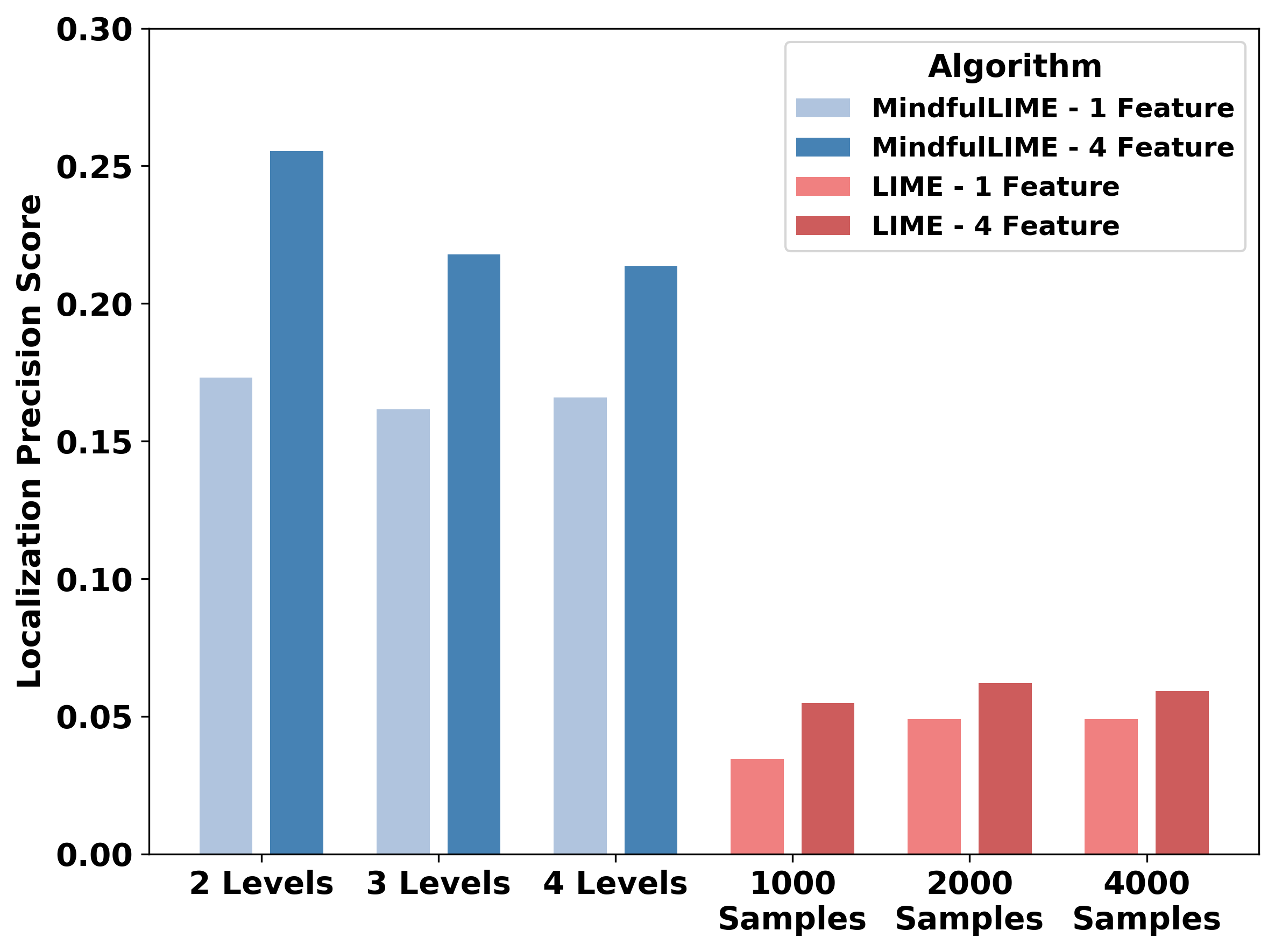}\\
  (c) Felzenszwalb \\
\end{tabular}
\caption{Quality Comparison: LIME vs. MindfulLIME for various versions and settings}
\label{fig:8}       
\end{figure*}

Moving on to Figure \ref{fig:8}, it presents the localization similarity between the ground truth annotations and the explanations generated by different variations of LIME and MindfulLIME, considering various segmentation algorithms and feature settings. A higher localization precision score indicates greater consistency between the explanation generated by an algorithm and the actual annotated data in terms of bounding box distance. Since each sample instance in our data can contain multiple regions marked by doctors for the same detected condition, we conducted experiments with the top 1 feature and the top 4 features to observe the effect of increasing the number of top superpixels in the generated explanations.

In all settings, MindfulLIME demonstrates higher localization precision or, in other words, higher explanation quality than LIME. Except for the Quickshif algorithm with the top 1 feature setting, MindfulLIME (2 Levels) consistently produces the best results in all other cases. Additionally, it is evident that increasing the number of top features to 4 leads to an increase in localization precision for both algorithms, which can be attributed to the multi-label nature of samples and the granularity of the marked annotations in actual data.

\begin{figure*}
    \centering
    \includegraphics[width=0.60\textwidth]{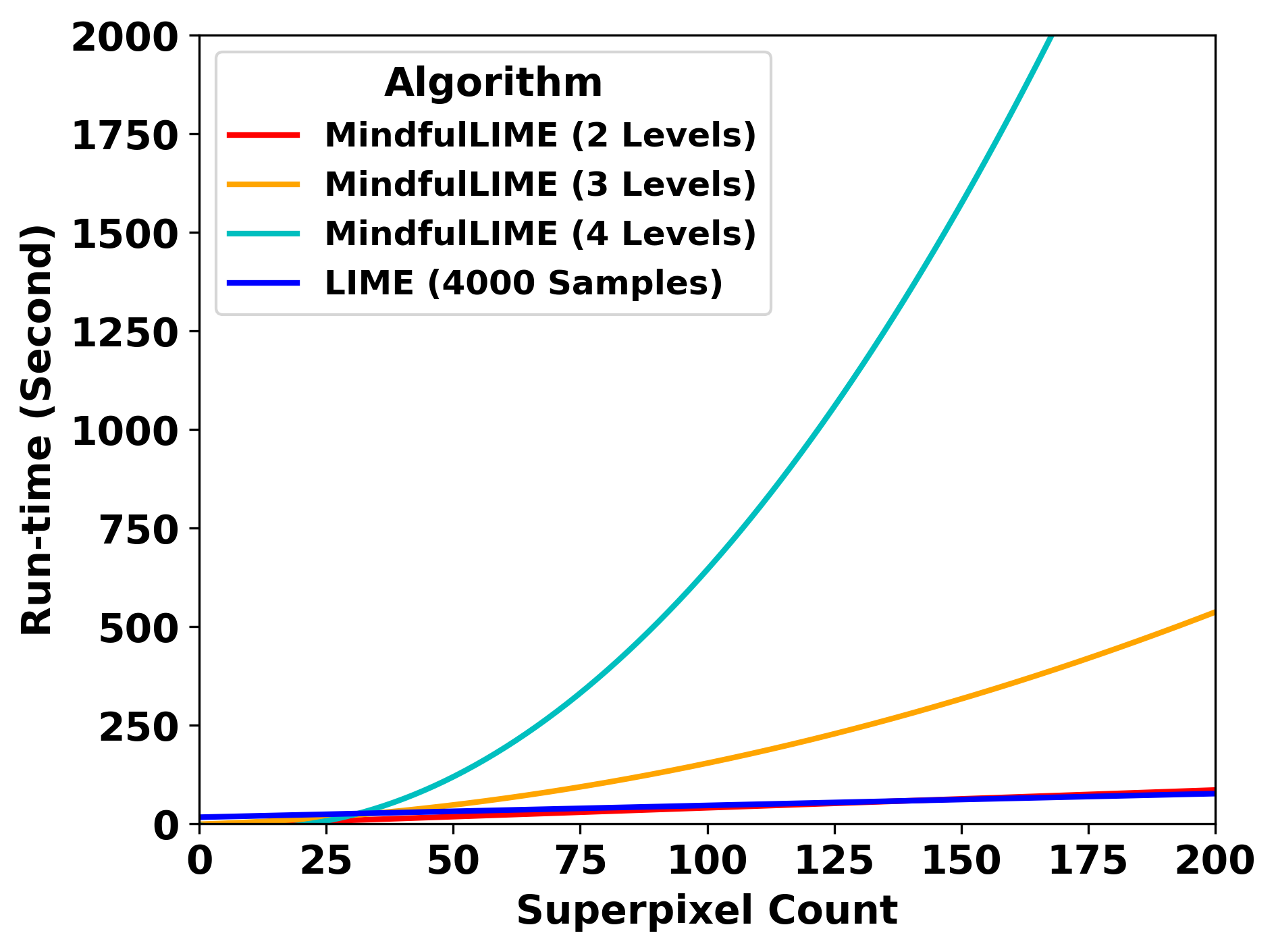}
    \caption{Run-Time Comparison: LIME vs. MindfulLIME for various versions and superpixel numbers}\label{fig:9}
\end{figure*}

The run-time of LIME and MindfulLIME is compared with different numbers of superpixels generated by various segmentation methods, as shown in Figure \ref{fig:9} and Table \ref{table:2}. Each segmentation algorithm produces varying numbers, sizes, and arrangements of segments within an image. For instance, as shown in Table \ref{table:2}, SLIC divides images into 49 superpixels, Quickshift into 66, and Felzenswalb into 158 on average. 

While LIME's run-time primarily depends on the number of perturbed samples, MindfulLIME's complexity is influenced by the number of superpixels (which affect the maximum adjacent superpixels) and the algorithm's depth level. As the depth level increases in MindfulLIME, the impact of the number of superpixels, especially the maximum adjacent superpixels, intensifies. This is due to the exponential increase in explored branches and subsequent processes. The effect of this intensification is demonstrated in Table \ref{table:2}, where superpixel algorithms that produce denser and more segments, such as Felzenswalb, require additional computations and lead to longer processing times.

MindfulLIME proves to be a practical choice, achieving a favorable balance between stability, precision, and efficiency. It demonstrates these qualities without needing deeper levels or intricate computations, particularly by employing less complex superpixel algorithms like SLIC and Quickshift with a higher granularity of segments. Our findings highlight the superiority of MindfulLIME over LIME in various segmentation settings, as it generates fewer high-quality samples within an acceptable processing time. Overall, MindfulLIME guarantees stability and improves localization precision, making it a valuable tool for generating reliable and accurate explanations in machine learning models.

\begin{table}
\centering
\caption{Run-Time Comparison: LIME vs. MindfulLIME for various versions and superpixel numbers}\label{table:2}
\begin{tabular}{@{}lllll@{}}
\toprule
Algorithm & Version & Average Superpixel & Average Run-time & Average Generated  \\
          &         & Count              & (Second)         & Samples \\
\midrule
LIME            &   1000 Samples   & 49     & 11.6        & 1000                            \\
                &                  & 66     & 12.0        & 1000                            \\
                &                  & 158    & 13.8        & 1000                            \\
                &   2000 Samples   & 49     & 23.2         & 2000                            \\
                &                  & 66     & 23.9        & 2000                            \\
                &                  & 158    & 27.6         & 2000                            \\
                &   4000 Samples   & 49     & 46.2        & 4000                            \\
                &                  & 66     & 47.7        & 4000                            \\
                &                  & 158    & 54.9        & 4000                            \\
MindfulLIME     &   2 Levels       & 49     & 14.5         & 173                           \\
                &                  & 66     & 21.7         & 261                             \\
                &                  & 158    & 69.4        & 834                          \\
                &   3 Levels       & 49     & 40.5        & 513                          \\
                &                  & 66     & 79.8        & 1016                         \\
                &                  & 158    & 349.0       & 4196                            \\
                &   4 Levels       & 49     & 91.7        & 1183                         \\
                &                  & 66     & 258.0       & 3275                         \\
                &                  & 158    & 1763.0      & 17836                        \\
\end{tabular}
\end{table}

\section{Conclusions and Future Work}\label{section:6}

In summary, MindfulLIME is engineered to optimize both stability and performance without compromising efficiency. It proposes a purposive sample generation algorithm to eradicate randomness, employs a graph-based algorithm to consider correlated features, and applies uncertainty sampling techniques to precisely localize explanations. Together, these components address the instability issues associated with existing models like LIME which using random perturbation sampling, guaranteeing the production of more reliable and interpretable explanations for black-box models.

Despite advancements in the stability of other data modalities, a notable gap remains for image data. Our rigorous training and evaluation on the CheXpert and VinDr-CXR datasets, respectively, confirm that MindfulLIME not only attains 100\% stability but also efficiently surpasses LIME in localization precision while generating fewer samples. This achievement demonstrates MindfulLIME's potential to enhance the interpretability of black-box models, particularly in critical domains. These extensive experiments not only bridge this gap but also underline the model's potential for broader applications beyond the medical imaging domain addressed in this work. The selection of these datasets implies that MindfulLIME can be effectively generalized across other domains, given the challenges associated with interpreting such images. By overcoming the fundamental limitations of LIME, MindfulLIME ensures stable and accurate explanations, thereby amplifying the interpretability of black-box models. This positions MindfulLIME as a valuable, durable, and indispensable tool for the future.

While MindfulLIME has achieved stability and shown promising results, future research could explore enhancements to its performance and efficiency. A significant area for potential development involves refining the sample filtering process. This could include both investigating alternative techniques and integrating high-confidence samples to more effectively discern informative samples and identify Out-of-Distribution data. Such refinements could optimize graph pruning policies and potentially balance the depth of explanations with reliability, especially in handling borderline cases. Additionally, the exploration of various feature selection methods is another avenue that might contribute to more precise and interpretable explanations. While currently focused on medical images, MindfulLIME's adaptable design suggests its applicability to a wider range of graph-convertible data samples. Future research might extend the application of MindfulLIME to diverse datasets and domains to assess its generalizability and effectiveness in different contexts. These directions represent possible pathways for enhancing MindfulLIME, aimed at further optimizing its performance across various applications.

\section*{Statements and Declarations}

\textbf{Competing Interests} All authors certify that they have no affiliations with or involvement in any organization or entity with any financial or non-financial interest in the subject matter or materials discussed in this manuscript. This research received no specific grant from funding agencies in the public, commercial, or not-for-profit sectors.\\

\noindent
\textbf{Data Availability} The CheXpert dataset \cite{article} analyzed during the current study is a public chest X-ray dataset available at \url{https://stanfordaimi.azurewebsites.net/datasets/8cbd9ed4-2eb9-4565-affc-111cf4f7ebe2} upon request and with the permission and under the license of Stanford University Dataset Research Use Agreement. Similarly, the VinDr-CXR dataset \cite{Nguyen2020} that supports the findings of this study can be accessed through \url{https://physionet.org/content/vindr-cxr/1.0.0/} in case of fulfilling the requirements. 

%\bibliographystyle{unsrt}  
%\bibliography{main_article}

\end{document}